\pdfoutput=1

\documentclass[11pt]{article}

\usepackage{emnlp2021}

\usepackage{times}
\usepackage{latexsym}
\usepackage{todonotes}
\usepackage[T1]{fontenc}
\usepackage{url}

\usepackage[utf8]{inputenc}

\usepackage{microtype}
\usepackage{booktabs}
\usepackage{listings}
\title{CUNI systems for WMT21: Terminology translation Shared Task}

\author{Josef Jon \and Michal Novák \and João Paulo Aires \and Dušan Variš \and Ondřej Bojar \\
       Charles University\\
  \texttt{\{jon,aires,varis,bojar\}@ufal.mff.cuni.cz}  }

\begin{document}
\maketitle
\begin{abstract}
This  paper  describes  Charles  University  submission for Terminology translation Shared Task at WMT21.  The objective of this task is to design a system which translates certain terms based on a provided terminology database, while preserving high overall translation quality. We competed in English-French language pair. Our approach is based on providing the desired translations alongside the input sentence and  training the model to use these provided terms. We lemmatize the terms both during the training and inference, to allow the model to learn how to produce correct surface forms of the words, when they differ from the forms provided in the terminology database. Our submission ranked second in Exact Match metric which evaluates the ability of the model to produce desired terms in the translation.
\end{abstract}

\section{Introduction}
Terminology integration, or, more generally, constrained translation in NMT was extensively studied in recent years. Lexically constrained translation means that aside from the source sentence, we have available some additional knowledge of what tokens or expressions should appear in the translation and we want to force the system to include them in the generated output. Three main ways of enforcing these constraints have been studied.

First, replacing the source part of the constraint that is found in the source sentence with a placeholder which is then copied by the model into the output. There it gets replaced by the target part of the constraint (\citet{luong-etal-2015-addressing}; \citet{crego2016systrans}).

Second way is to modify the decoding search algorithm in a way that only allows hypotheses containing the constraints to be marked as finished (\citet{anderson-etal-2017-guided}; \citet{hasler-etal-2018-neural}; \citet{chatterjee-etal-2017-guiding}; \citet{hokamp-liu-2017-lexically}; \citet{post-vilar-2018-fast}; \citet{hu-etal-2019-improved})

Finally, some works focus on providing the constraints directly to the model as part of the input sequence. The model is trained to incorporate these constraints into the output, for example \citet{dinu-etal-2019-training}; 
\citet{chen-etal-2020-lexical}; 
\citet{song-etal-2019-code} or \citet{bergmanis-pinnis-2021-facilitating}. 

As apparent from previous paragraphs, the problem of integrating lexical constraints into NMT is well studied, but one issue was largely ignored. In inflected languages, the  surface form of the constraint in the output cannot be known beforehand, as there are usually many possible ways to translate a sentence and many of them need different surface forms of the constraint to be fluent and grammatically correct. For example, let's say we have a terminology database containing term pair \textit{influenza -> grippe} and this source sentence:
\begin{quote}
    During the 2018-2019 \textbf{influenza} season.
\end{quote}
Possible correct translation is:
\begin{quote}

Pendant la saison \textbf{grippale}  2018-2019.
\end{quote}
Where the term base noun form \textit{grippe} is inflected into adjective \textit{grippale}. Traditional constraint integration methods will try to enforce the term DB form \textit{grippe} instead.

We have studied this problem in our recent work \cite{jon-etal-2021-end} concurrently with \citet{bergmanis-pinnis-2021-facilitating}, who used a very similar approach. Both works use different languages and evaluation pipelines and both show that the proposed approach is feasible.

\section{Method}
NMT models are known to produce fluent, consistent and grammatically correct outputs \cite{popel-etal-2020-cubbitt}. Thus, it makes sense to utilize this ability of the model to inflect the constraint into correct form, instead of trying to disambiguate the form externally.
Our approach is based on annotating source sentences with the desired target constraints and training the model to incorporate these constraints into the output. We publish our preprocessing scripts at \url{https://github.com/ufal/bergamot/wmt21-terminology}

\subsection{Term annotation}
There are multiple possibilities in how to exactly annotate the source sentence. For example, let's say the terminology database contains entries: \begin{quote}
\textit{runny nose -> nez qui coule } \\
\textit{fever -> fièvre }

\end{quote}

and we have a sentence:
\begin{quote}
    
\textit{And are you having a \textbf{runny nose} or \textbf{fever}?}
\end{quote}

One way is to replace the part of the source sentence containing the source constraint with the target part of the constraint:
\begin{quote}
    
\textit{And are you having a \textbf{nez qui coule} or \textbf{fièvre}?}
\end{quote}

Another option is to insert the translation tokens after the source part of the constraint and use factors to mark which tokens of a sentence belong to source constraint, which tokens are part of the target constraint and which are neither. For example, if factor with value 2 denotes that the token is part of the translation, value 1 means that the token is part of a source constraint and 0 means that it is just ordinary token, we get:
\begin{quote}

\textit{And$_0$ are$_0$ you$_0$ having$_0$ a$_0$ \textbf{runny$_1$ nose$_1$} \textbf{nez$_2$ qui$_2$ coule$_2$} or$_0$  \textbf{fever$_1$ } \textbf{fièvre$_2$ } ?$_0$}
\end{quote}

We use simpler method to integrate the constraints in our systems: we append them to the source sentence as a suffix, separated by a special token (\texttt{<sep>}) and in case of multiple constraints for a single sentence, we separate them by a different token (\texttt{<c>}):
\begin{quote}
    
\textit{And are you having a \textbf{runny nose} or \textbf{fever}? <sep>  \textbf{nez qui coule} <c> \textbf{fièvre}}
\end{quote}

For more details about the possible modifications of our method, comparisons with other approaches and detailed evaluation and analysis, we refer the reader to our previous work \cite{jon-etal-2021-end}. 

\subsection{Training data generation}
We prepare synthetic constraints for parallel training data by sampling random token subsequences from the target sentence. These subsequences are used as a suffix for the source sentence as described earlier. There is a number of parameters guiding this process. Every token in a sentence can become a start of a constraint with probability $s$. Unless stated otherwise, we set $s=0.1$. Any subsequent token in an open constraint can end the constraint with probability $e=0.75$. We permit multiple non-overlapping constraints for a sentence. We skip the sentence for constraint generation (i.e. leave it without any constraints) with probability $n=0.1$. In pseudocode:

\begin{lstlisting}[language=Python,basicstyle=\small]
s=0.1
e=0.75
n=0.1
for sent in text:
  r=random()
  constraints=[]
  if r > n:
    open=False
    constraint=""
    for t in tokens(sent):
      r=random()
      if open:
        if  r < e:
          constraints.append(constraint)
          open=False
        else:
          constraint+=t
      else:
        if  r < s:
          constraint+=t
          open=True
  print(sent, constraints)

\end{lstlisting}
Since the task allows for multiple target variants for a single source term, we have to account for such possibility in our training data generation. We assume that each generated constraint can have a variant with probability $v=0.1$. This variant is sampled randomly (with no relation to the source sentence) from n-grams extracted from the  target training corpus (so it is not a part of a current target sentence, but it is still a plausible subsequence in the target language). The variant has the same number of tokens as the original constraint with probability $l=0.9$, otherwise the length of the variant is taken from triangular distribution between 1 and 9 with mode 2. The variants of a single constraint are delimeted with another special token \texttt{<v>}. None of the probabilities were tuned for improving results, we chose them based on manual inspection of the generated data. We use values that produced similar counts and lengths of the constraints as in the validation set.

\subsection{Lemmatization}

The training data generation method described above works, but suffers from the issues described in the introduction -- the system learns to generate only the exact tokens supplied as  constraints in the suffix, but doesn't account for different possible inflections of the constraints in different contexts. To overcome this issue, we lemmatize the constraints both during the training and during test time. This way, the model learns to not only generate the correct words in the output, but also to correctly inflect them.

\subsection{Source-side terminology matching}

To find term pairs from terminology database in the input text, we lemmatize both the database source side and input sentences and search for the terms that appear either on lemma or surface form level. Since our lemmatizer works with context, we lemmatize both the text and the database word by word to ensure consistent lemmas. 
For the models trained with lemmatized constraints, we lemmatize also the target side of the terminology database and anntote the source sentence with lemmas of the target terms, instead of the surface forms.

\section{Experiments}
\subsection{Data}
We used all English-French corpora allowed by the organizers, aside from Paracrawl (with the exception of one model, which is marked). Namely this means Europarl v10, Common Crawl, UN Parallel Corpus v1.0, News Commentary v16 and Gigaword. We used WMT15 news test set as our validation set. After deduplication and filtering, the resulting  training set consists of 24.6M sentences without Paracrawl and 125.9M including Paracrawl.

\subsection{Tools}
We use MarianNMT~\cite{junczys-dowmunt-etal-2018-marian-fast} to train Transformer-big models with standard parameters~\cite{vaswani-2017-attention}.
The corpora are filtered using Moses cleaning script\footnote{\url{https://github.com/marian-nmt/moses-scripts}} and \texttt{fasttext} langid \cite{joulin2016bag}. We split the text into subwords using FactoredSegmenter\footnote{\url{https://github.com/microsoft/factored-segmenter}} based on SentencePiece~\cite{kudo-richardson-2018-sentencepiece} and lemmatize using UDPipe~\cite{straka-strakova-2017-tokenizing}. BLEU scores are computed using SacreBLEU \cite{post-2018-call}, other metric are obtained by an evaluation script provided by the organizers\footnote{\url{https://github.com/mahfuzibnalam/terminology_evaluation}}\cite{alam2021evaluation}.

\begin{table*}[t]

\resizebox{\linewidth}{!}{
\begin{tabular}{llcccccc}
\textbf{Constraints} & \textbf{Corpus} & \multicolumn{1}{l}{\textbf{Variants}} & \multicolumn{1}{l}{\textbf{BLEU}} & \multicolumn{1}{l}{\textbf{EM}} & \multicolumn{1}{l}{\textbf{window 2}} & \multicolumn{1}{l}{\textbf{window 3}} & \multicolumn{1}{l}{\textbf{1-TERm}} \\ \toprule
None                 & Base            & -                                     & 43.976                            & 0.862                           & 0.289                                 & 0.283                                 & 0.584                               \\
None                 & Base+paracrawl  & -                                     & 45.084                            & 0.851                           & 0.283                                 & 0.279                                 & 0.587                               \\
None                 & Base+bt         & -                                     & 42.319                            & 0.834                           & 0.282                                 & 0.275                                 & 0.575                               \\
SF                   & Base            & no                                    & 43.771                            & 0.953                           & 0.297                                 & 0.290                                 & 0.581                               \\
SF                   & Base            & yes                                   & 41.656                            & 0.982                           & 0.253                                 & 0.255                                 & 0.555                               \\
Lemm                 & Base            & yes                                   & 42.317                            & 0.919                           & 0.278                                 & 0.274                                 & 0.552                               \\
Lemm                 & Base            & no                                    & 44.959                            & 0.961                           & 0.302                                 & 0.296                                 & 0.591                               \\
Lemm*                 & Base            & no                                    & 44.623	&0.909 &0.292&	0.288	&0.588                        \\

Final combined       & -               & -                                     & 45.590                            & 0.989                           & 0.309                                 & 0.304                                 & 0.600                              
\end{tabular}
}
\caption{Results of our models on official validation set. The first column specifies whether the constraint were lemmatized (\textit{Lemm}) or not SF (\textit{SF}), second one shows which part of copora we used. Base means all parallel data allowed by the organizers with exception of Paracrawl. Third column says whether we provided all possible variants of the target term from terminology database to the model, on we only the first one. Asterisk in \textit{Constraints} column means that the model was trained with these form of constraints, but no constraints were provided during the test time.} 
\label{tab:results}
\end{table*}
\subsection{Evaluation}
The script provided by the task organizers computes multiple metrics: BLEU, (Lemmatized) Exact Match, Window overlap and 1-TERm. 

Exact match is a fraction of constraints which were produced in the outputs (the output sentences are lemmatized and the search is performed on both lemma and surface form level). This metric can be cheated in two ways -- first, the system can place the target constraint at arbitrary place in the output, e.g. we can just translate with a non-constrained MT model, append the constraints at the end and obtain a perfect score. 
Second way is related to lemmatization -- the system can produce any valid surface form of the constraint and even though this form is not grammatically correct in context of the output sentence, it still gets counted as matching. On the other hand, without lemmatization, only the word forms listed in the terminology database would get accepted, which would not cover all the possible correct forms.

Window overlap aims to overcome the first shortcoming of EM by evaluating placement of the constraint in the output. For each constraint in the translation and in the reference, windows of \textit{n} tokens are extracted and compared with each other to see if the system places the constraint in similar context as in the reference.  2 and 3 token windows are used.

TERm metric is weighted TER which uses higher weights for tokens which are part of a term from terminology database to increase sensitivity to differences in the terminology. In the experiments, we observed that 1-TERm score is influenced mainly by the overall translation quality and less so by the term integration. We believe that this metric alone is also not sufficient for comparing ability to  integrate constraints in different models, as the results seem to rely mainly on the "baseline" model performance, i.e. big general NMT model, trained on more data, which provides better overall translation quality, but does not explicitly integrate constraints, may obtain higher scores than a smaller constrained model with perfect constraint integration ability.

\subsection{Results}

We trained our models by techniques described earlier and we present metrics computed by the official evaluation script in Table \ref{tab:results}.  Due to time and computing constraints, most of the models were trained without Paracrawl corpus and we only trained one baseline on dataset including Paracrawl for comparison. We compared integrating constraints in the surface form (so the model needs to produce exactly the same token as provided in the input) and constraints in lemmatized form (the model can produce different inflection of the provided constraint). We also compared providing all possible variants of the target constraint from terminology database (delimeted by \texttt{<v>}, as described earlier), or just the first possible translation.  

We see that in most metrics, the model which is trained with lemmatized constraints and uses only one variant performs the best. Systems trained with multiple variants of the target term show large degradation in BLEU scores. We suppose one of the problems in our method is that during training, only the true constraint variant from the target is plausible translation of the source, others are n-grams sampled randomly from the whole corpus. Thus, the negative samples are very easy to distinguish during the training, but during the test time, the variants are provided by the term base and they are all plausible in the context. We will analyse these results further in the future.

Our final primary submission is a combination of all the models. They are ranked by their respective BLEU scores on validation set and we check if the produced translation contains the desired term either at lemma level. We use the best ranking systems' translation that does, or, in case none of the systems produced the term, we use the translation of baseline system.

The task organizers provide test set results.%
\footnote{\scriptsize{\url{https://docs.google.com/spreadsheets/d/13-lkwDq9yerehSF4No6ZTLqPXjSaL7HOsksnZDjjO-Y/}}}
Two metrics were considered for the ranking. First, COMET \cite{rei-etal-2020-comet}, which evaluates general translation quality without special regard for specific terminology. Secondly, exact match, which measures how many of the desired constraints were actually produced in the output, but suffers from the issues described earlier. Our primary submission was ranked on joint 6th-10th place out of 21 systems according to COMET and 1st-3rd according to exact match.

\subsection{Error analysis}

\begin{table*}[t]
\small

\resizebox{\linewidth}{!}{
\begin{tabular}{lp{0.39\linewidth}p{0.10\linewidth}p{0.41\linewidth}}
\textbf{i} &   \textbf{Source}  & \textbf{Target terms}     & \textbf{MT output}         \\                                                                                                \toprule

1 & Many human \textbf{Coronavirus} have their origin in bats. & coronavirus & Beaucoup de \textbf{Coronavirus} humains ont leur origine dans les chauves-souris .\\ \\
2 & Data from these practices are reported online in a weekly return, which includes monitoring weekly rates of influenza-like illness (ILI) and other communicable and \textbf{respiratory diseases} in England. & maladies respiratoires / maladies communes des voies respiratoires / maladie respiratoire & Les données relatives à ces pratiques sont communiquées en ligne dans une déclaration hebdomadaire, qui comprend le suivi des taux hebdomadaires de maladies grippales(SG) et d'autres \textbf{maladies} transmissibles et \textbf{respiratoires} en Angleterre. \\ \\
3-4 & We will share the protocol with UK colleagues and the I-MOVE consortium who have recently obtained EU Horizon 2020 funding from the stream “Advancing knowledge for the clinical and public health response to the \textbf{novel coronavirus epidemic}” &  coronavirus nouveau; epidémie / épidémies / épidémique   & Nous partagerons le protocole avec nos collègues du Royaume-Uni et le consortium I-MOVE , qui ont récemment obtenu un financement de l’OMS horizon 2020 dans le cadre du projet «Advancing knowledge for the clinical and public health response to the \textbf{novel coronavirus epidemic}»  \\
5 & The statistical methodology is in support of a policy approach to widespread disease \textbf{outbreak}, where so-called nonpharmaceutical interventions (NPIs) are used to respond to an emerging pandemic to produce disease suppression. & épidémie / épidémies / épidémique & La méthodologie statistique est à l'appui d'une approche politique face à l'apparition de\textbf{ maladies à grande échelle}, où les interventions dites non pharmaceutiques (ISP) sont utilisées pour répondre à une pandémie émergente afin d'éliminer les maladies. \\

\\

\end{tabular}
}
\caption{Rest of the examples with uncovered terms. \textit{Target terms} column shows possible translations of the source terms (bold) as provided in the terminology database.}
\label{tab:errors}
\end{table*}
Our submitted system did not cover 10 out of 872 term occurrences in the validation set. We analyse these ten errors manually. 
Six of these errors are related to casing, notably by translating \textit{SARS-CoV} as \textit{Sars-CoV}, instead of keeping the original casing (five occurrences). This is caused by our lemmatization pipeline, which produces \textit{Sars}  as lemma of \textit{SARS}. We confirmed that after manually fixing the input and restoring the original casing, the system produces correct output.
Other five examples classified as errors are presented in  Table \ref{tab:errors}.

Another casing error occurs in translation of the sentence (1) in the table. The model keeps the original source casing, but the evaluation script only checks for lower-case \textit{coronavirus}. This sentence is also actually part of unsplit and wrongly tokenized source line \textit{The large number of host bat and avian species, and their global range, has enabled extensive evolution and dissemination of coronaviruses.Many human coronavirus have their origin in bats.} This may be a source of further confusion for the model.

In example (2), the related terminology DB pair is \textit{respiratory diseases -> maladies respiratoires}. In the model output, the adjective \textit{transmissibles} is interjected between the terms, which is probably not an error from human point of view, but is hard to evaluate automatically.

In example (3-4), the model does not translate the name of the project in quotes, thus it does not produce the desired translations of  both \textit{epidemic -> épidemie} and \textit{novel coronavirus -> coronavirus nouveau }.

Finally, (5) is a true failure of the model to use the provided term. The sentence produced by the model is a plausible and semantically correct translation, but it is not using the desired term. For further analysis, we manually replaced the produced translation of the term (\textit{maladies à grande échelle}) with the term from the terminology database (\textit{épidémie}). We computed cross-entropy scores for the modified sentence both with and without providing the constraint to the model. We saw that when provided with the constraint, the modified translation is more probable than without the constraint (but still slightly less probable than the translation that was actually produced.) This shows that the method still partially works in this case, but the bias towards producing the term in the output needs to be stronger -- we plan to explore this further using contrastive learning.
\section{Conclusion}
We describe our submission to Terminology translation Shared Task at WMT21. We show our method can effectively incorporate the terminology without negative effects on overall translation quality. We analysed all ten examples in the validation set where our model did not cover the desired term constraint and we show that most of them can be  explained by preprocessing issues.
\section*{Acknowledgements}
Our work is supported by the Bergamot project (European Union’s Horizon 2020 research and innovation programme under grant agreement No 825303) aiming for fast and private user-side browser translation, GA ČR NEUREM3 grant (Neural Representations in Multi-modal and Multi-lingual Modelling, 19-26934X (RIV: GX19-26934X)) and by SVV 260 575 grant.

The work described herein has also been using data provided by the LINDAT/CLARIAH-CZ Research Infrastructure, supported by the Ministry of Education, Youth and Sports of the Czech Republic (Project No. LM2018101).
\bibliography{custom}

\begin{thebibliography}{22}
\expandafter\ifx\csname natexlab\endcsname\relax\def\natexlab#1{#1}\fi

\bibitem[{Anderson et~al.(2017)Anderson, Fernando, Johnson, and
  Gould}]{anderson-etal-2017-guided}
Peter Anderson, Basura Fernando, Mark Johnson, and Stephen Gould. 2017.
\newblock \href {https://doi.org/10.18653/v1/D17-1098} {Guided open vocabulary
  image captioning with constrained beam search}.
\newblock In \emph{Proceedings of the 2017 Conference on Empirical Methods in
  Natural Language Processing}, pages 936--945, Copenhagen, Denmark.
  Association for Computational Linguistics.

\bibitem[{Bergmanis and Pinnis(2021)}]{bergmanis-pinnis-2021-facilitating}
Toms Bergmanis and M{\=a}rcis Pinnis. 2021.
\newblock \href {https://www.aclweb.org/anthology/2021.eacl-main.271}
  {Facilitating terminology translation with target lemma annotations}.
\newblock In \emph{Proceedings of the 16th Conference of the European Chapter
  of the Association for Computational Linguistics: Main Volume}, pages
  3105--3111, Online. Association for Computational Linguistics.

\bibitem[{Chatterjee et~al.(2017)Chatterjee, Negri, Turchi, Federico, Specia,
  and Blain}]{chatterjee-etal-2017-guiding}
Rajen Chatterjee, Matteo Negri, Marco Turchi, Marcello Federico, Lucia Specia,
  and Fr{\'e}d{\'e}ric Blain. 2017.
\newblock \href {https://doi.org/10.18653/v1/W17-4716} {Guiding neural machine
  translation decoding with external knowledge}.
\newblock In \emph{Proceedings of the Second Conference on Machine
  Translation}, pages 157--168, Copenhagen, Denmark. Association for
  Computational Linguistics.

\bibitem[{Chen et~al.(2020)Chen, Chen, Wang, and Li}]{chen-etal-2020-lexical}
Guanhua Chen, Yun Chen, Yong Wang, and Victor~O.K. Li. 2020.
\newblock \href {https://doi.org/10.24963/ijcai.2020/496}
  {Lexical-constraint-aware neural machine translation via data augmentation}.
\newblock In \emph{Proceedings of the Twenty-Ninth International Joint
  Conference on Artificial Intelligence, {IJCAI-20}}, pages 3587--3593.
  International Joint Conferences on Artificial Intelligence Organization.
\newblock Main track.

\bibitem[{Crego et~al.(2016)Crego, Kim, Klein, Rebollo, Yang, Senellart,
  Akhanov, Brunelle, Coquard, Deng, Enoue, Geiss, Johanson, Khalsa, Khiari, Ko,
  Kobus, Lorieux, Martins, Nguyen, Priori, Riccardi, Segal, Servan, Tiquet,
  Wang, Yang, Zhang, Zhou, and Zoldan}]{crego2016systrans}
Josep Crego, Jungi Kim, Guillaume Klein, Anabel Rebollo, Kathy Yang, Jean
  Senellart, Egor Akhanov, Patrice Brunelle, Aurelien Coquard, Yongchao Deng,
  Satoshi Enoue, Chiyo Geiss, Joshua Johanson, Ardas Khalsa, Raoum Khiari,
  Byeongil Ko, Catherine Kobus, Jean Lorieux, Leidiana Martins, Dang-Chuan
  Nguyen, Alexandra Priori, Thomas Riccardi, Natalia Segal, Christophe Servan,
  Cyril Tiquet, Bo~Wang, Jin Yang, Dakun Zhang, Jing Zhou, and Peter Zoldan.
  2016.
\newblock \href {http://arxiv.org/abs/1610.05540} {Systran's pure neural
  machine translation systems}.

\bibitem[{Dinu et~al.(2019)Dinu, Mathur, Federico, and
  Al-Onaizan}]{dinu-etal-2019-training}
Georgiana Dinu, Prashant Mathur, Marcello Federico, and Yaser Al-Onaizan. 2019.
\newblock \href {https://doi.org/10.18653/v1/P19-1294} {Training neural machine
  translation to apply terminology constraints}.
\newblock In \emph{Proceedings of the 57th Annual Meeting of the Association
  for Computational Linguistics}, pages 3063--3068, Florence, Italy.
  Association for Computational Linguistics.

\bibitem[{Hasler et~al.(2018)Hasler, de~Gispert, Iglesias, and
  Byrne}]{hasler-etal-2018-neural}
Eva Hasler, Adri{\`a} de~Gispert, Gonzalo Iglesias, and Bill Byrne. 2018.
\newblock \href {https://doi.org/10.18653/v1/N18-2081} {Neural machine
  translation decoding with terminology constraints}.
\newblock In \emph{Proceedings of the 2018 Conference of the North {A}merican
  Chapter of the Association for Computational Linguistics: Human Language
  Technologies, Volume 2 (Short Papers)}, pages 506--512, New Orleans,
  Louisiana. Association for Computational Linguistics.

\bibitem[{Hokamp and Liu(2017)}]{hokamp-liu-2017-lexically}
Chris Hokamp and Qun Liu. 2017.
\newblock \href {https://doi.org/10.18653/v1/P17-1141} {Lexically constrained
  decoding for sequence generation using grid beam search}.
\newblock In \emph{Proceedings of the 55th Annual Meeting of the Association
  for Computational Linguistics (Volume 1: Long Papers)}, pages 1535--1546,
  Vancouver, Canada. Association for Computational Linguistics.

\bibitem[{Hu et~al.(2019)Hu, Khayrallah, Culkin, Xia, Chen, Post, and
  Van~Durme}]{hu-etal-2019-improved}
J.~Edward Hu, Huda Khayrallah, Ryan Culkin, Patrick Xia, Tongfei Chen, Matt
  Post, and Benjamin Van~Durme. 2019.
\newblock \href {https://doi.org/10.18653/v1/N19-1090} {Improved lexically
  constrained decoding for translation and monolingual rewriting}.
\newblock In \emph{Proceedings of the 2019 Conference of the North {A}merican
  Chapter of the Association for Computational Linguistics: Human Language
  Technologies, Volume 1 (Long and Short Papers)}, pages 839--850, Minneapolis,
  Minnesota. Association for Computational Linguistics.

\bibitem[{ibn Alam et~al.(2021)ibn Alam, Anastasopoulos, Besacier, Cross,
  Gallé, Koehn, and Nikoulina}]{alam2021evaluation}
Md~Mahfuz ibn Alam, Antonios Anastasopoulos, Laurent Besacier, James Cross,
  Matthias Gallé, Philipp Koehn, and Vassilina Nikoulina. 2021.
\newblock \href {http://arxiv.org/abs/2106.11891} {On the evaluation of machine
  translation for terminology consistency}.

\bibitem[{Jon et~al.(2021)Jon, Aires, Varis, and Bojar}]{jon-etal-2021-end}
Josef Jon, Jo{\~a}o~Paulo Aires, Dusan Varis, and Ond{\v{r}}ej Bojar. 2021.
\newblock \href {https://doi.org/10.18653/v1/2021.acl-long.311} {End-to-end
  lexically constrained machine translation for morphologically rich
  languages}.
\newblock In \emph{Proceedings of the 59th Annual Meeting of the Association
  for Computational Linguistics and the 11th International Joint Conference on
  Natural Language Processing (Volume 1: Long Papers)}, pages 4019--4033,
  Online. Association for Computational Linguistics.

\bibitem[{Joulin et~al.(2016)Joulin, Grave, Bojanowski, and
  Mikolov}]{joulin2016bag}
Armand Joulin, Edouard Grave, Piotr Bojanowski, and Tomas Mikolov. 2016.
\newblock Bag of tricks for efficient text classification.
\newblock \emph{arXiv preprint arXiv:1607.01759}.

\bibitem[{Junczys-Dowmunt et~al.(2018)Junczys-Dowmunt, Grundkiewicz, Dwojak,
  Hoang, Heafield, Neckermann, Seide, Germann, Fikri~Aji, Bogoychev, Martins,
  and Birch}]{junczys-dowmunt-etal-2018-marian-fast}
Marcin Junczys-Dowmunt, Roman Grundkiewicz, Tomasz Dwojak, Hieu Hoang, Kenneth
  Heafield, Tom Neckermann, Frank Seide, Ulrich Germann, Alham Fikri~Aji,
  Nikolay Bogoychev, Andr\'{e} F.~T. Martins, and Alexandra Birch. 2018.
\newblock \href {http://www.aclweb.org/anthology/P18-4020} {Marian: Fast neural
  machine translation in {C++}}.
\newblock In \emph{Proceedings of ACL 2018, System Demonstrations}, pages
  116--121, Melbourne, Australia. Association for Computational Linguistics.

\bibitem[{Kudo and Richardson(2018)}]{kudo-richardson-2018-sentencepiece}
Taku Kudo and John Richardson. 2018.
\newblock \href {https://doi.org/10.18653/v1/D18-2012} {{S}entence{P}iece: A
  simple and language independent subword tokenizer and detokenizer for neural
  text processing}.
\newblock In \emph{Proceedings of the 2018 Conference on Empirical Methods in
  Natural Language Processing: System Demonstrations}, pages 66--71, Brussels,
  Belgium. Association for Computational Linguistics.

\bibitem[{Luong et~al.(2015)Luong, Sutskever, Le, Vinyals, and
  Zaremba}]{luong-etal-2015-addressing}
Thang Luong, Ilya Sutskever, Quoc Le, Oriol Vinyals, and Wojciech Zaremba.
  2015.
\newblock \href {https://doi.org/10.3115/v1/P15-1002} {Addressing the rare word
  problem in neural machine translation}.
\newblock In \emph{Proceedings of the 53rd Annual Meeting of the Association
  for Computational Linguistics and the 7th International Joint Conference on
  Natural Language Processing (Volume 1: Long Papers)}, pages 11--19, Beijing,
  China. Association for Computational Linguistics.

\bibitem[{Popel et~al.(2020)Popel, Tomkova, Tomek, Łukasz Kaiser, Uszkoreit,
  Bojar, and {\v{Z}}abokrtsk{\'{y}}}]{popel-etal-2020-cubbitt}
Martin Popel, Marketa Tomkova, Jakub Tomek, Łukasz Kaiser, Jakob Uszkoreit,
  Ond{\v{r}}ej Bojar, and Zden{\v{e}}k {\v{Z}}abokrtsk{\'{y}}. 2020.
\newblock Transforming machine translation: a deep learning system reaches news
  translation quality comparable to human professionals.
\newblock \emph{Nature Communications}, 11(4381):1--15.

\bibitem[{Post(2018)}]{post-2018-call}
Matt Post. 2018.
\newblock \href {https://doi.org/10.18653/v1/W18-6319} {A call for clarity in
  reporting {BLEU} scores}.
\newblock In \emph{Proceedings of the Third Conference on Machine Translation:
  Research Papers}, pages 186--191, Brussels, Belgium. Association for
  Computational Linguistics.

\bibitem[{Post and Vilar(2018)}]{post-vilar-2018-fast}
Matt Post and David Vilar. 2018.
\newblock \href {https://doi.org/10.18653/v1/N18-1119} {Fast lexically
  constrained decoding with dynamic beam allocation for neural machine
  translation}.
\newblock In \emph{Proceedings of the 2018 Conference of the North {A}merican
  Chapter of the Association for Computational Linguistics: Human Language
  Technologies, Volume 1 (Long Papers)}, pages 1314--1324, New Orleans,
  Louisiana. Association for Computational Linguistics.

\bibitem[{Rei et~al.(2020)Rei, Stewart, Farinha, and
  Lavie}]{rei-etal-2020-comet}
Ricardo Rei, Craig Stewart, Ana~C Farinha, and Alon Lavie. 2020.
\newblock \href {https://doi.org/10.18653/v1/2020.emnlp-main.213} {{COMET}: A
  neural framework for {MT} evaluation}.
\newblock In \emph{Proceedings of the 2020 Conference on Empirical Methods in
  Natural Language Processing (EMNLP)}, pages 2685--2702, Online. Association
  for Computational Linguistics.

\bibitem[{Song et~al.(2019)Song, Zhang, Yu, Luo, Wang, and
  Zhang}]{song-etal-2019-code}
Kai Song, Yue Zhang, Heng Yu, Weihua Luo, Kun Wang, and Min Zhang. 2019.
\newblock \href {https://doi.org/10.18653/v1/N19-1044} {Code-switching for
  enhancing {NMT} with pre-specified translation}.
\newblock In \emph{Proceedings of the 2019 Conference of the North {A}merican
  Chapter of the Association for Computational Linguistics: Human Language
  Technologies, Volume 1 (Long and Short Papers)}, pages 449--459, Minneapolis,
  Minnesota. Association for Computational Linguistics.

\bibitem[{Straka and Strakov{\'a}(2017)}]{straka-strakova-2017-tokenizing}
Milan Straka and Jana Strakov{\'a}. 2017.
\newblock \href {https://doi.org/10.18653/v1/K17-3009} {Tokenizing, {POS}
  tagging, lemmatizing and parsing {UD} 2.0 with {UDP}ipe}.
\newblock In \emph{Proceedings of the {C}o{NLL} 2017 Shared Task: Multilingual
  Parsing from Raw Text to Universal Dependencies}, pages 88--99, Vancouver,
  Canada. Association for Computational Linguistics.

\bibitem[{Vaswani et~al.(2017)Vaswani, Shazeer, Parmar, Uszkoreit, Jones,
  Gomez, Kaiser, and Polosukhin}]{vaswani-2017-attention}
Ashish Vaswani, Noam Shazeer, Niki Parmar, Jakob Uszkoreit, Llion Jones,
  Aidan~N. Gomez, undefinedukasz Kaiser, and Illia Polosukhin. 2017.
\newblock Attention is all you need.
\newblock In \emph{Proceedings of the 31st International Conference on Neural
  Information Processing Systems}, NIPS'17, page 6000–6010, Red Hook, NY,
  USA. Curran Associates Inc.

\end{thebibliography}

\bibliographystyle{acl_natbib}

\end{document}